\documentclass[conference]{IEEEtran}
\IEEEoverridecommandlockouts
\usepackage{cite}
\usepackage{amsmath,amssymb,amsfonts}
\usepackage{algorithmic}
\usepackage{graphicx}
\usepackage{subcaption}
\usepackage{textcomp}
\usepackage{xcolor}
\usepackage{siunitx}
\usepackage{soul}
\def\BibTeX{{\rm B\kern-.05em{\sc i\kern-.025em b}\kern-.08em
    T\kern-.1667em\lower.7ex\hbox{E}\kern-.125emX}}
\begin{document}

\title{Augmenting Automated Game Testing with Deep Reinforcement Learning
}

\author{\IEEEauthorblockN{Joakim Bergdahl, Camilo Gordillo, Konrad Tollmar, Linus Gisslén}
\IEEEauthorblockA{
\textit{SEED - Electronic Arts (EA)}, Stockholm, Sweden \\
jbergdahl, cgordillo, ktollmar, lgisslen@ea.com}
}
\maketitle

\begin{abstract}
General game testing relies on the use of human play testers, play test scripting, and prior knowledge of areas of interest to produce relevant test data. Using deep reinforcement learning (DRL), we introduce a self-learning mechanism to the game testing framework. With DRL, the framework is capable of exploring and/or exploiting the game mechanics based on a user-defined, reinforcing reward signal. As a result, test coverage is increased and unintended game play mechanics, exploits and bugs are discovered in a multitude of game types. In this paper, we show that DRL can be used to increase test coverage, find exploits, test map difficulty, and to detect common problems that arise in the testing of first-person shooter (FPS) games.

\end{abstract}


\section{Introduction}
When creating modern games, hundreds of developers, designers and artists are often involved. Game assets amount to thousands, map sizes are measured in square kilometers, and characters and abilities are often abundant. As games become more complex, so do the requirements for testing them, thus increasing the need for automated solutions where the sole use of human testers is both impractical and expensive. Common methods involve scripting behaviours of classical in-game AI actors offering scalable, predictable and efficient ways of automating testing. However, these methods present drawbacks when it comes to adaptability and learnability. See section \ref{sec:method} for further discussion. 

Reinforcement learning (RL) models open up the possibility of complementing current scripted and automated solutions by learning directly from playing the game without the need of human intervention. Modern RL algorithms are able to explore complex environments \cite{mirowski2016learning} while also being able to find exploits in the game mechanics \cite{Baker2020Emergent}. RL fits particularly well in modern FPS games which, arguably, consist of two main phases: navigation (finding objectives, enemies, weapons, health, etc.) and combat (shooting, reloading, taking cover, etc.). Recent techniques have tackled these kinds of scenarios using either a single model learning the dynamics of the whole game \cite{harmer18}, or two models focusing on specific domains respectively (navigation and combat) \cite{lample2017playing}. 

\section{Previous Work}
Research in game testing has provided arguments for automated testing using Monte-Carlo simulation and handcrafted player models to play and predict the level of difficulty in novel parameter configurations of a given game \cite{isaksen2015exploring}. Supervised learning using human game play data has successfully been applied to testing in 2D mobile puzzle games by predicting human-like actions to evaluate the playability of game levels \cite{gudmundsson2018human}. Reinforcement learning has also been used for quality assurance where a Q-learning based model is trained to explore graphical user interfaces of mobile applications \cite{koroglu2018qbe}. Applications of reinforcement learning in game testing show the usability of human-like agents for game evaluation and balancing purposes as well as the problematic nature of applying the technology in a game production setting \cite{borovikov2019winning}. Active learning techniques have been applied in games to decrease the amount of human play testing needed for quality assurance by preemptively finding closer-to-optimal game parameters ahead of testing \cite{zook2019automatic}. 
In the work that is most similar to ours, a combination of evolutionary algorithms, DRL and multi-objective optimization is used to test online combat games \cite{zheng2019wuji}. However, in this paper we take a modular approach where RL is used to complement classical test scripting rather than replace it. We expand on this research including exploit detection and continuous actions (simulating continuous controllers like mouse and game-pads) while focusing on navigation (as opposed to combat).




\section{Method}
\label{sec:method}
The methodology presented in this paper is the result from close collaboration between researchers and game testers.

\subsection{Reinforcement Learning}
Scripted agents are appealing due to their predictable and reproducible behaviours. However, scripting alone is not a perfect solution and there are key motivations for complementing it with RL.

\begin{itemize}
    \item RL agents have the capacity to learn from interactions with the game environment \cite{harmer18}, as opposed to traditional methods. This results in behaviours more closely resembling those of human players, thus increasing the probability of finding bugs and exploits.
    \item Scripting and coding behaviours can be both hard and cumbersome. Moreover, any change or update to the game is likely to require rewriting or updating existing test scripts. RL agents, on the contrary, can be retrained or fine-tuned with minimal to no changes to the general setup. RL agents are also likely to learn complex policies which would otherwise remain out of reach for classical scripting.
    \item RL agents can be controlled via the reward signal to express a specific behaviour. Rewards can be used to encourage the agents to play in a certain style, e.g. to make an agent play more defensively one can alter the reward function towards giving higher rewards for defensive behaviours. The reward signal can also be used to imitate human players and to encourage exploration by means of curiosity \cite{pathak2017curiosity}.
\end{itemize}

Both scripted and RL agents are scalable in the sense that it is possible to parallelize and execute thousands of agents on a few machines. With these properties, we argue that RL is an ideal technique for augmenting automated testing and complementing classical scripting. 

\subsection{Agent Controllers}
In scripted automatic testing it is a common practice to use pre-baked navigation meshes to allow agents to move along the shortest paths available. Navigation meshes, however, are not designed to resemble the freedom of movement that a human player experiences. Any agent exclusively following these trajectories will fail to explore the environment to the same degree a human would and it is therefore likely to miss navigation-related bugs.

In the following experiments, the RL agents use continuous controller inputs corresponding to a game controller, i.e. forward/backward, left/right turn, left/right strafe, and jump. No navigation meshes are introduced in the demonstrated environments.  


\subsection{Agent observation and reward function}
The observation state for all agents in this paper is an aggregated observation vector consisting of: agents position relative to the goal ($\mathbb{R}^{3}$), agents velocity ($\mathbb{R}^{3}$), agent world rotation ($\mathbb{R}^{4}$), goal distance ($\mathbb{R}$), is climbing ($\mathbb{B}$), contact with ground ($\mathbb{B}$), jump cool-down time ($\mathbb{R}$), reset timer ($\mathbb{R}$) and a vision array. The vision array consists of 12 ray casts in various directions.  All values are normalized to be kept between $[-1, 1]$. The agents receive an incremental, positive reward for moving towards a goal and an equally sized negative reward as a penalty for moving away from it. 

\subsection{Environments}

\begin{figure}[!t]
    \centering
    \includegraphics[width=0.40\textwidth]{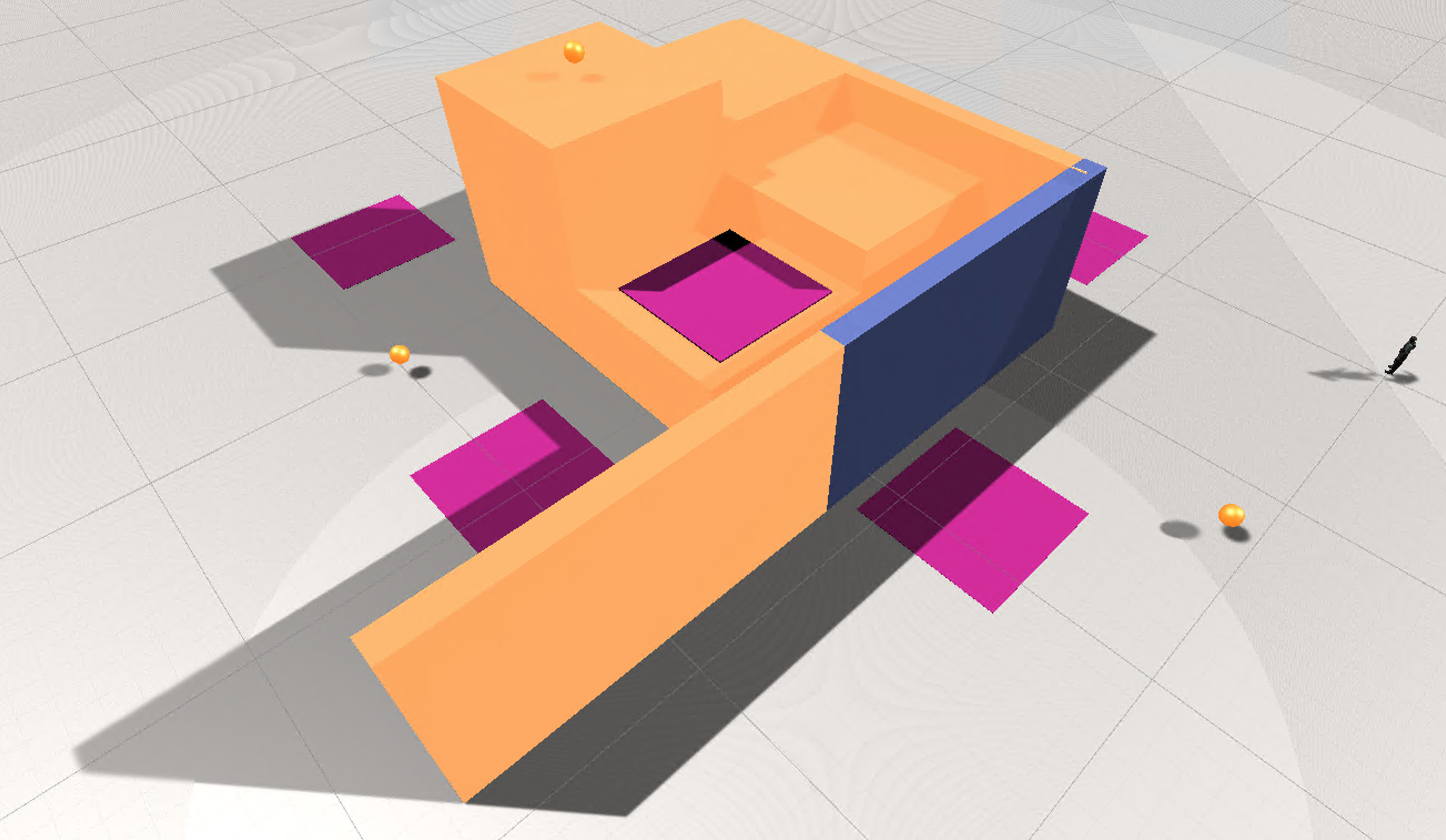}
    \caption{{\it Exploit} and {\it Stuck Player} sand-box: The yellow spheres indicate navigation goals. The blue wall lacks a collision mesh which allows agents to walk through it. The pink squares represent areas where the agent will get stuck when entering them.}
    \label{fig:ExploitEnv}
\end{figure}

\begin{figure}[!t]
    \centering
    \includegraphics[width=0.40\textwidth]{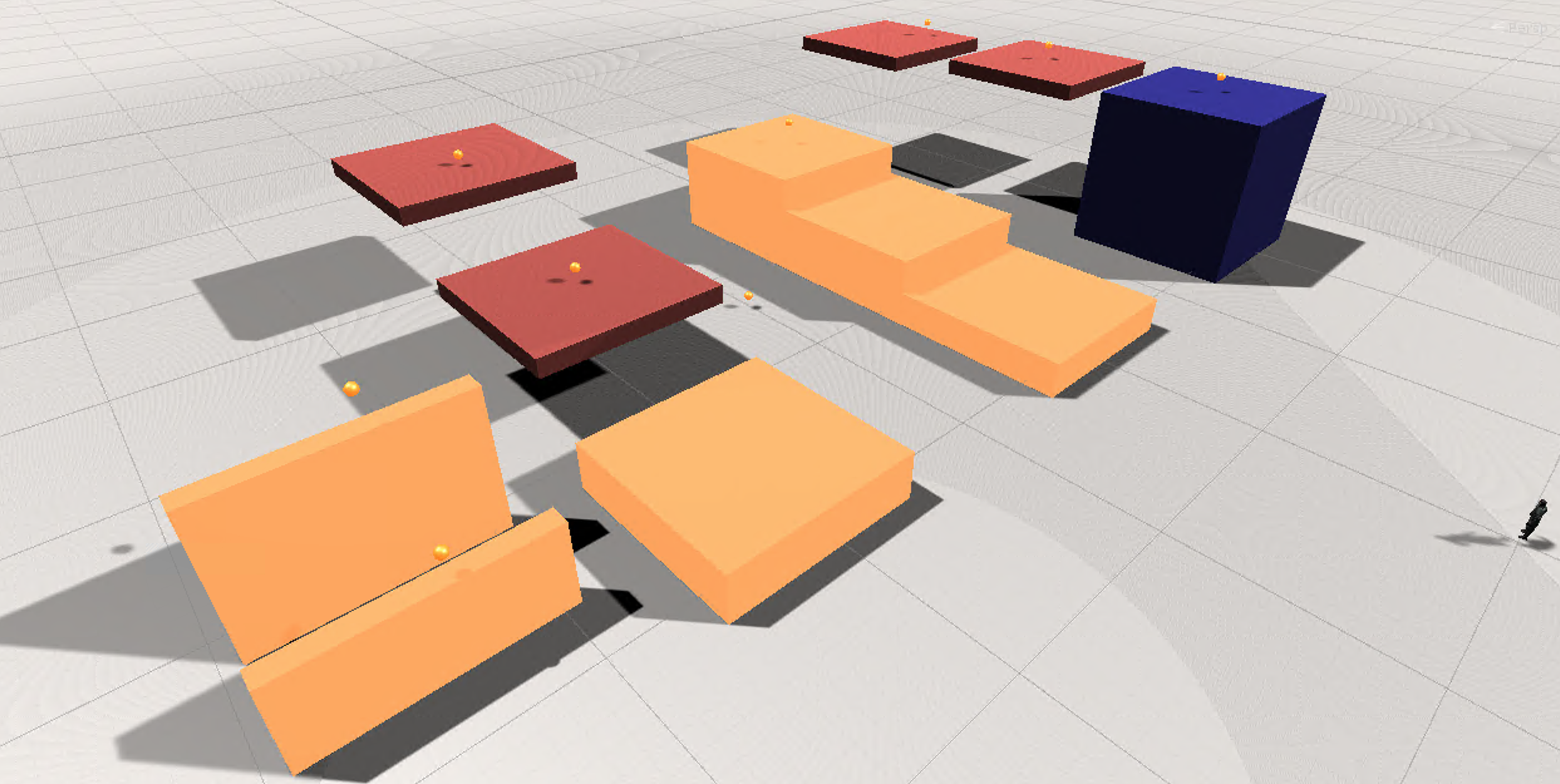}
    \caption{{\it Navigation} and {\it Dynamic Navigation} sand-box: The yellow spheres indicate navigation goals. The dark blue box represents a climbable wall which is the only way to reach the two goals farthest away from the camera. In the {\it Dynamic Navigation} environment the 4 red platforms move.}
    \label{fig:NavigationEnv}
\end{figure}

We test our hypothesis on a set of sand-box environments where we can test certain bug classes (e.g. navigation bugs, exploits, etc.). We investigate local navigation tasks by letting the environments represent a smaller part of larger maps where hundreds of agents could be deployed to test different areas. The agents always start in the same positions in the environment so as not to use random placement as a mean to explore the map. We employ four different sand-box environments:

\begin{itemize}
    \item {\it Exploit} - One of the walls in the environment lacks a collision mesh thus allowing the agent to exploit a short-cut that is not intentional. See Fig. \ref{fig:ExploitEnv}.
    \item {\it Stuck Player} - This environment is similar to the {\it Exploit} sand-box but with five areas where the agent will get stuck when entering. The goal here is to identify all these areas. See Fig. \ref{fig:ExploitEnv}. 
    \item {\it Navigation} - This  environment is based on a complex navigation task. The task is to reach navigation goals in the shortest time possible. It requires the agent to jump and climb in order to reach the goals. See Fig. \ref{fig:NavigationEnv}.
    \item {\it Dynamic Navigation} - This environment is identical to the Navigation sand-box but with moving, traversable platforms in the scene. See Fig. \ref{fig:NavigationEnv}.
\end{itemize}

\subsection{Test scenarios}

This paper focuses on navigation in FPS type games. However, the approach we use is transferable not only to other elements of an FPS game such as target selection and shooting but also to other types of games. We apply RL to a set of different areas: {\it game exploits and bugs}, {\it distribution of visited states}, and {\it difficulty evaluation}.

\subsection{Training}
\begin{figure}[!t]
    \centering
    \includegraphics[width=0.50\textwidth]{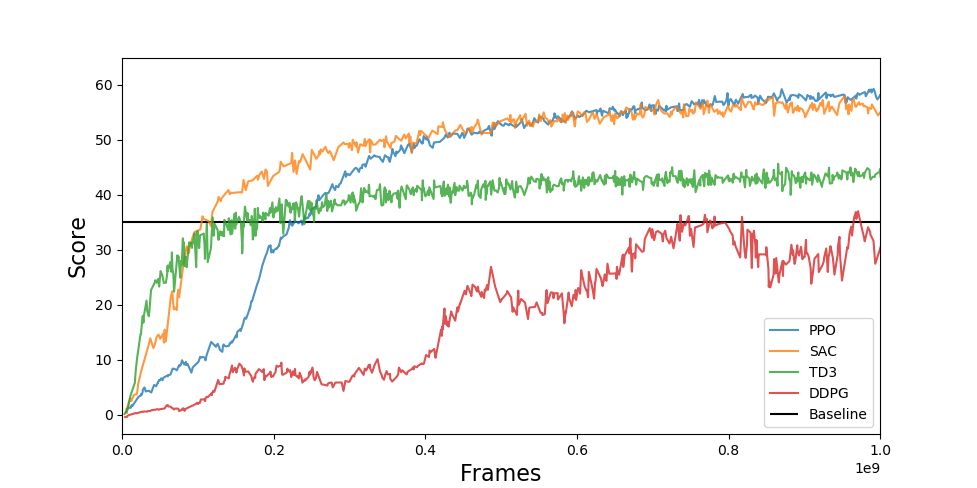}
    \caption{Comparison between different algorithms training on the {\it Dynamic Navigation} sand-box. All models were trained using the same learning rate. Baseline is a human player.}
    \label{fig:Results}
\end{figure}

We compare different algorithms (see Fig. \ref{fig:Results}) and for the experiments in this paper, Proximal Policy Optimization (PPO) tends to reach higher scores without a significant increase in training time \cite{schulman2017proximal}. For this reason, we will be using PPO and report its performance in the following. We train the models using a training server hosted on one machine (AMD Ryzen Threadripper 1950X @ 3.4 GHz,  Nvidia GTX 1080 Ti) being served data from clients running on 4 separate machines (same CPUs as above).
During evaluation, the algorithms interact with the environment at an average rate of 10000 interactions/second. With an action repeat of 3, this results in a total of 30000 actions/second. Assuming a human is able to perform 10 in-game actions per second, the performance average over the algorithms corresponds to 3000 human players. With this setup, only a fraction of the machines and time required for a corresponding human play test is needed.


Training the agents in the various environments requires between 50 - 1000 M environment frames depending on the complexity of the game. Fig. \ref{fig:Results} shows a detailed comparison between the different algorithms for the {\it Dynamic Navigation} environment. All agents in the other environments and tests are trained with identical algorithm (i.e. PPO), hyper-parameters, observations, and reward function, i.e. they are not tailored to the task at hand. In the following section we present our findings and discuss them in detail.

\section{Results}
\label{sec:results}


\subsection{Game exploits and logical bugs}
A game exploit is a class of bugs that gives the player the means to play the game with an unfair advantage. Exploits may lead to unbalanced game play ultimately resulting in a deteriorated player experience. Examples of game exploits are the possibility of moving through walls or hide inside objects (making the player invisible to others) due to missing collision meshes. In comparison, logical game bugs introduce unpredictable and unintended effects to the game play experience. One example are those areas in a level where the player gets stuck and cannot leave. In the worst case scenario, logical bugs may render the game unplayable.

One of the main differences between traditional game AI (i.e. scripting) and machine learning is the ability of the latter to learn from playing the game. This ability allows RL agents to improve as they interact with the game and learn the underlying reward function. Moreover, there are plenty of examples where RL algorithms have found unintended mechanics in complex scenarios \cite{Baker2020Emergent}. Unlike human play testers, however, RL agents have no prior understanding of how a game is intended to be played. Although this could be regarded as a disadvantage, in the context of game testing, we argue, it becomes a useful attribute for finding exploits.

\begin{figure}[!t]
    \centering
    \begin{subfigure}[b]{0.22\textwidth}  
        \centering 
        \includegraphics[width=\textwidth, trim={5cm 8cm 8cm 5cm},clip] {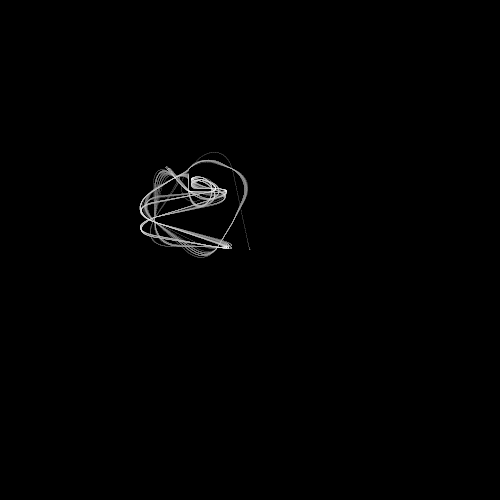}
        \caption{Scripted NavMesh agent.}    
        \label{fig:ScriptedAgentExploitEnv}
    \end{subfigure}
    \hfill
    \begin{subfigure}[b]{0.22\textwidth}
        \centering
        \includegraphics[width=\textwidth, trim={5cm 8cm 8cm 5cm},clip] {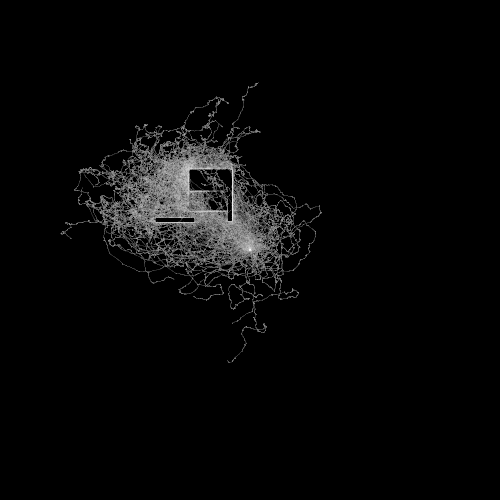}
        \caption{RL agent after 5 M steps.}    
        \label{fig:RLAgentExpEnv01}
    \end{subfigure}
    
    \vskip\baselineskip
    \begin{subfigure}[b]{0.22\textwidth}
        \centering
        \includegraphics[width=\textwidth, trim={5cm 8cm 8cm 5cm},clip] {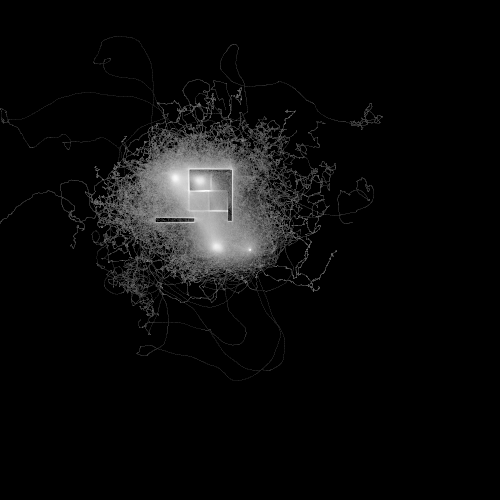}
        \caption{RL agent after 30 M steps.}    
        \label{fig:RLAgentExpEnv02}
    \end{subfigure}
    \hfill
    \begin{subfigure}[b]{0.22\textwidth}  
        \centering 
        \includegraphics[width=\textwidth,trim={5cm 8cm 8cm 5cm},clip] {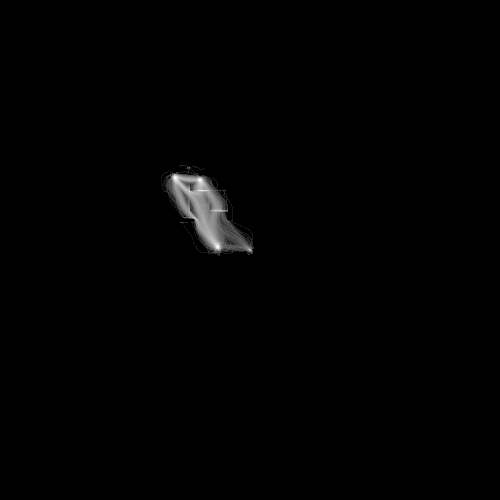}
        \caption{RL agent fully trained.}    
        \label{fig:RLAgentExpEnv03}
    \end{subfigure}
    
    \caption{Heat maps generated during training on the {\it Exploit} sand-box, see Fig. \ref{fig:ExploitEnv}. Fig. \ref{fig:ScriptedAgentExploitEnv} shows the scripted agent following its navigation system output. Figs. (b), (c), and (d) show how the distribution of the agents changes during training. We see that early in training the visited states are evenly distributed across the map. When fully trained, the agents find the near optimal path to the goals.}
    \label{fig:my_label}
\end{figure}

We use the {\it Exploit} environment (see Fig. \ref{fig:ExploitEnv}) as an example of what could happen when the lack of a collision mesh opens up a short-cut in the environment. Comparing Figs. \ref{fig:ScriptedAgentExploitEnv} and \ref{fig:RLAgentExpEnv03} we see how the two agent types (scripted and RL) behave. In Fig. \ref{fig:ScriptedAgentExploitEnv} it is evident that the scripted navigation mesh agent is unaware of the available short-cut. In contrast, the RL agent quickly finds a way of exploiting the missing collision mesh and learns to walk through the wall to maximize its reward (see Fig. \ref{fig:RLAgentExpEnv03}).

\begin{figure}[!t]
    \centering
    \includegraphics[width=0.40\textwidth]{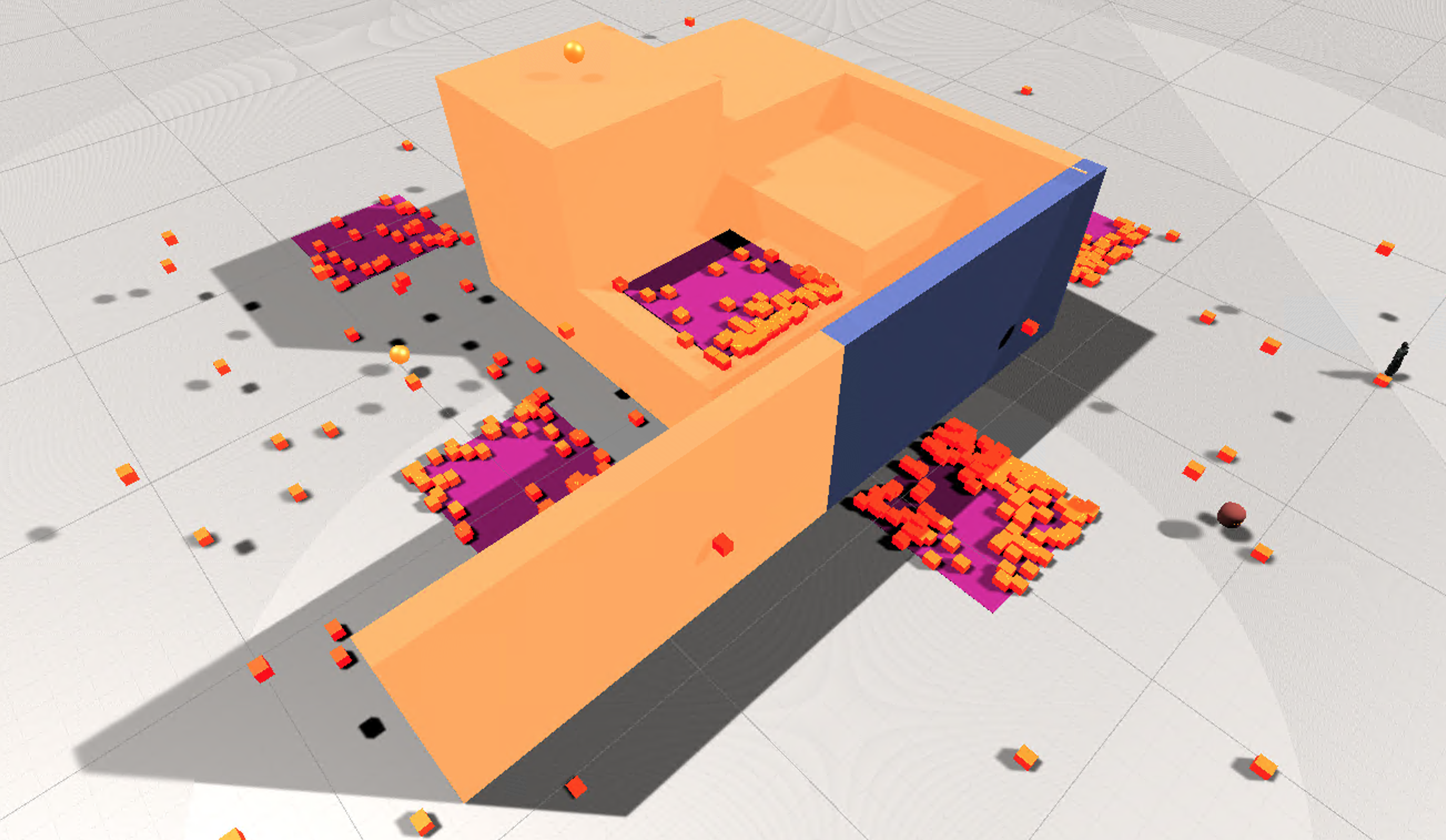}
    \caption{Results from {\it Stuck Player} sand-box: The small boxes show where the agents timeout, clearly indicating where on the map the agents get stuck.}
    \label{fig:StuckEnvTimeout}
\end{figure}

Furthermore, using a navigation mesh is not an effective way of finding areas where players could get stuck. These areas are often found in places human players are not expected to reach and are therefore very rarely covered by the navigation mesh. In the following experiment we used the {\it Stuck Player} sand-box to analyze the positions in the map where agents would time-out. Fig. \ref{fig:StuckEnvTimeout} shows the positions of the agents at the end of each training episode. From visual inspection alone it is clear that all five areas of interest could be identified.

\subsection{Distribution of game states visited}
\label{sec:state_distribution}
\begin{figure}[!t]
    \centering
    \begin{subfigure}[b]{0.22\textwidth}
        \centering
        \includegraphics[width=\textwidth, trim={5cm 8cm 6.5cm 3.5cm},clip] {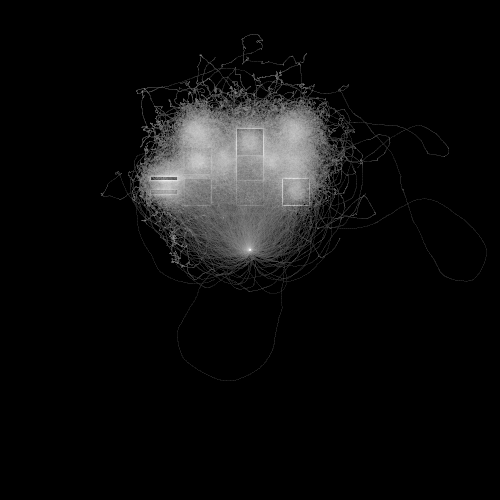}
        \caption{RL agent after 20 M steps.}    
        \label{fig:RLAgentDynEnv01}
    \end{subfigure}%
    \hfill
    \begin{subfigure}[b]{0.22\textwidth}  
        \centering 
        \includegraphics[width=\textwidth,trim={5cm 8cm 6.5cm 3.5cm},clip] {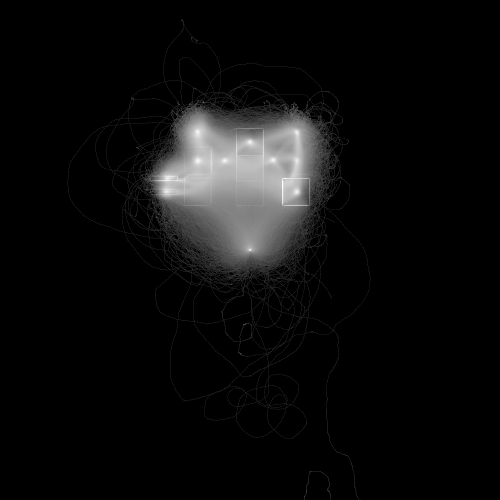}
        \caption{RL agent fully trained.}    
        \label{fig:RLAgentDynEnv03}
    \end{subfigure}
    
    \caption{Heat maps generated during training on the {\it Dynamic Navigation} sand-box. See Fig. \ref{fig:NavigationEnv}. The white dots indicate that the agent managed to reach a navigation goal. One of the goals that is harder to reach can be seen to the left in the heat maps. Reaching this goal requires the agent to jump on two thin ledges which only occurs at the end of  training (Fig. (b)).}
    \label{fig:RLAgentDynEnv}
\end{figure}

Adequate test coverage is important to successfully test server load, graphical issues, performance bottlenecks, etc. In order to maximize the test coverage it is desirable to visit all reachable states within the game. Human testers naturally achieve this as they generally play with diverse objective functions and play styles. Traditional testing using navigation meshes leads to a mechanical and repeating pattern, see for example Fig. \ref{fig:ScriptedAgentExploitEnv}. In the RL approach, the agents frequently update their behaviours during training ranging from exploration to exploit focused (compare Figs. \ref{fig:RLAgentDynEnv01} and \ref{fig:RLAgentDynEnv03}).

\subsection{Difficulty evaluation}

The time it takes for the agent to master a task can be used as an indicator of how difficult the game would be for a human player. Table \ref{tab:difficult} shows a comparison of the number of frames required to train navigation agents in different environments. As the complexity of the tasks increases (due to larger and/or dynamic scenarios), so does the time required to train the agents. We envision these metrics being used to measure and compare the difficulty of games.

\begin{table}[!t]
\centering
\begin{tabular}{lll}
\hline
{\it Sand-box}               & 80\% of max    & 50\% of max  \\ \hline
{\it Exploit-Explore}        & 88.2 M         & 70.8 M   \\
{\it Navigation}             & 298.8 M         & 197.4 M    \\
{\it Dynamic Navigation}     & 342.0 M         & 224.1 M   \\
\hline
\end{tabular}
\caption{Comparison of identical tasks but in different environments, see Figs. \ref{fig:ExploitEnv} and \ref{fig:NavigationEnv}. We report frames required to reach a certain percentage of max reward.}
\label{tab:difficult}
\end{table}

\section{Conclusion}
\label{sec:conclusion}
In this paper we have shown how RL can be used to augment traditional scripting methods to test video games. From experience in production, we have observed that RL is better suited for modular integration where it can complement rather than replace existing techniques. Not all problems are better solved with RL and training is substantially easier when focusing on single, well isolated tasks. RL can complement scripted tests in edge cases where human-like navigation, exploration, exploit detection and difficulty evaluation is hard to achieve.




\bibliography{refs}
\bibliographystyle{IEEEtran}

\end{document}